\documentclass{article}
\PassOptionsToPackage{numbers, compress}{natbib}

\usepackage[preprint]{neurips_2024}

\usepackage[utf8]{inputenc} 
\usepackage[T1]{fontenc}    
\usepackage{hyperref}       
\usepackage{url}            
\usepackage{booktabs}       
\usepackage{amsfonts}       
\usepackage{nicefrac}       
\usepackage{microtype}      
\usepackage{xcolor}         
\usepackage{graphicx}  
\usepackage{subcaption} 
\usepackage{multirow}
\usepackage{wrapfig}
\newcommand\mamba{\raisebox{-4pt}{\includegraphics[width=1.1em]{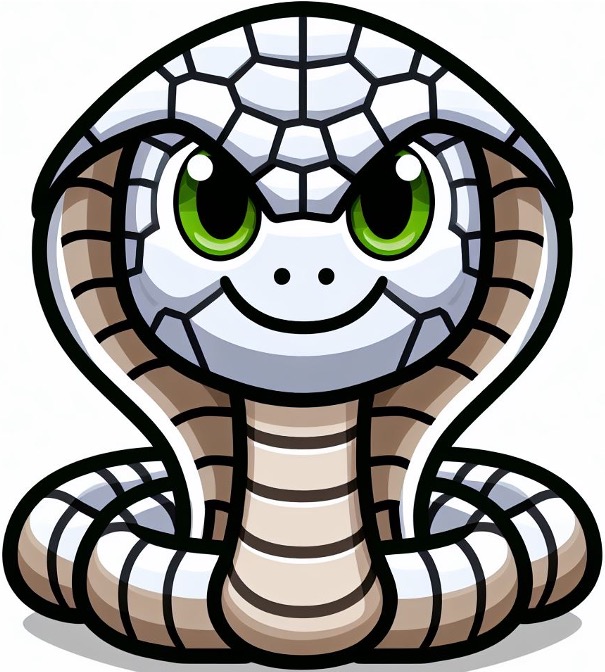}}}

\usepackage{amsmath,amsfonts,bm}

\def\eqref#1{equation~\ref{#1}}

\def\1{\bm{1}}

\def\vg{{\bm{g}}}
\def\vh{{\bm{h}}}

\def\vp{{\bm{p}}}

\def\vu{{\bm{u}}}
\def\vv{{\bm{v}}}
\def\vw{{\bm{w}}}
\def\vx{{\bm{x}}}
\def\vy{{\bm{y}}}

\def\mA{{\bm{A}}}
\def\mB{{\bm{B}}}
\def\mC{{\bm{C}}}
\def\mD{{\bm{D}}}

\def\mF{{\bm{F}}}

\def\mH{{\bm{H}}}
\def\mI{{\bm{I}}}

\def\mK{{\bm{K}}}

\def\mQ{{\bm{Q}}}

\def\mV{{\bm{V}}}
\def\mW{{\bm{W}}}

\DeclareMathAlphabet{\mathsfit}{\encodingdefault}{\sfdefault}{m}{sl}
\SetMathAlphabet{\mathsfit}{bold}{\encodingdefault}{\sfdefault}{bx}{n}

\newcommand{\R}{\mathbb{R}}

\newcommand{\softmax}{\mathrm{softmax}}

\DeclareMathOperator*{\patch}{PatchEmbed}
\DeclareMathOperator*{\mix}{Mix-SSM-Block}
\DeclareMathOperator*{\merging}{PatchMerging}
\DeclareMathOperator*{\globalpool}{GlobalPool}
\DeclareMathOperator*{\linear}{Linear}
\DeclareMathOperator*{\Conv}{Conv}
\DeclareMathOperator*{\SSM}{SSM}
\DeclareMathOperator*{\MLP}{MLP}
\DeclareMathOperator*{\MSA}{MSA}
\DeclareMathOperator*{\Attn}{Attn}
\DeclareMathOperator*{\Act}{Activation}
\DeclareMathOperator*{\CSM}{CSM}
\DeclareMathOperator*{\Order}{Order}
\DeclareMathOperator*{\Traverse}{Traverse}

\title{\mamba{}  InsectMamba: Insect Pest Classification with \\State Space Model}

\author{
  Qianning Wang$^{1}$, Chenglin Wang$^{2}$, Zhixin Lai$^{3}$, Yucheng Zhou$^{4}$ \\
  $^{1}$Nanjing Audit University, $^{2}$East China Normal University\\
  $^{3}$Snapchat, $^{4}$SKL-IOTSC, CIS, University of Macau \\
  \texttt{yucheng.zhou@connect.um.edu.mo} \\
}

\begin{document}

\maketitle

\begin{abstract}
The classification of insect pests is a critical task in agricultural technology, vital for ensuring food security and environmental sustainability. However, the complexity of pest identification, due to factors like high camouflage and species diversity, poses significant obstacles. Existing methods struggle with the fine-grained feature extraction needed to distinguish between closely related pest species. Although recent advancements have utilized modified network structures and combined deep learning approaches to improve accuracy, challenges persist due to the similarity between pests and their surroundings. To address this problem, we introduce InsectMamba, a novel approach that integrates State Space Models (SSMs), Convolutional Neural Networks (CNNs), Multi-Head Self-Attention mechanism (MSA), and Multilayer Perceptrons (MLPs) within Mix-SSM blocks. This integration facilitates the extraction of comprehensive visual features by leveraging the strengths of each encoding strategy. A selective module is also proposed to adaptively aggregate these features, enhancing the model's ability to discern pest characteristics. InsectMamba was evaluated against strong competitors across five insect pest classification datasets. The results demonstrate its superior performance and verify the significance of each model component by an ablation study.
\end{abstract}

\section{Introduction}
In agricultural production, due to pests significantly impacting crop yields, the identification and classification of pests within agricultural technology are pivotal for ensuring food security and sustainability. The insect pest classification aims to leverage vision models to automate insect pest recognition \cite{ForestryPestIdentification,IP102}. This task is crucial for maintaining crop health, potentially reducing pesticide usage, and fostering environmentally sustainable agricultural practices. Furthermore, accurate identification of insect pests benefits crop management by minimizing damage and optimizing yields.

Since pests often exhibit a high degree of camouflage within their natural habitats \cite{an2023insect,IP102}, which makes visual recognition difficult. This challenge also shows the complexity of insect pest classification. The similarity between pests and their surroundings, coupled with the vast diversity of species, poses significant obstacles to traditional image processing algorithms. Furthermore, the necessity for fine-grained feature extraction to distinguish between closely related pest species adds another layer of complexity to this challenge \cite{anwar2023exploring,butera2021precise}. Recent work has proposed utilizing modified capsule networks to improve network structure, thereby enhancing the hierarchical and spatial relationships of features to increase classification accuracy \cite{ForestryPestIdentification,butera2021precise}. Additionally, some studies have combined multiple deep networks and the advantages of complementary features from multiple perspectives to enhance recognition rates and robustness \cite{an2023insect,anwar2023exploring}. Nonetheless, these approaches still face challenges due to the similarity between pests and their surroundings.

In addressing the challenges of accurately identifying and classifying pests in varied conditions, different visual encoding strategies offer different advantages. Convolutional Neural Networks (CNNs \cite{CNN}) excel in local feature extraction, whereas the Multi-Head Self-Attention mechanism (MSA \cite{Transformer}) is adept at capturing global features. The State Space Models (SSMs \cite{Mamba}) structure is particularly effective at recognizing long-distance dependencies, and Multilayer Perceptrons (MLPs \cite{MLP}) specialize in channel-aware information inference.

To integrate advantages from different visual encoding strategies, we propose a novel approach, InsectMamba, consisting of Mix-SSM blocks that integrate SSM, CNN, MSA, and MLP to extract more comprehensive visual features for insect pest classification. In addition, we propose a selective module to adaptively aggregate visual features from different encoding strategies. Our method, leveraging the complementary capabilities of these visual encoding strategies, aims to achieve the vision model's capability in capturing both the local and global features of pests, thus addressing the critical challenges of camouflage and species diversity.

In the experiments, we evaluate our model and other strong competitors on five insect pest classification datasets. To improve the challenge of datasets, we re-split the dataset. The experimental results show that our method outperforms other methods, which demonstrates the effectiveness of our method. Moreover, we conduct the ablation study to verify the significance of each module of our model. Furthermore, we conduct extensive analysis of our model design to demonstrate its effectiveness.

Main contributions of this study are as follows:
\begin{itemize}
    \item We propose InsectMamba, which is the first attempt at the potential application of SSM-based models in insect pest classification.
    \item We present Mix-SSM blocks that seamlessly integrate SSM, CNN, MSA, and MLP. This integration allows our model to capture a comprehensive range of visual features for insect pest classification.
    \item We propose a selective aggregation module designed to adaptively combine visual features derived from different encoding strategies. This module allows the model to select relevant features that are utilized for classification.
    \item  We have rigorously evaluated InsectMamba across five insect pest classification datasets, demonstrating its superior performance compared to existing models.
\end{itemize}

\section{Related Work}
\subsection{Image Classification}
The rapid advancements in computer vision \cite{liu2024particle,zhou2024visual,su2024large,ZhangWHKR23} have led to its extensive application across various areas including AI security \cite{lyu2024task}, generated detection \cite{lai2024adaptive}, biomedicine \cite{lai2024language}, and agricultural technology \cite{wu2024new}. Notably, image classification \cite{Alexnet,ViT,VMamba} stands out as a fundamental technique for many applications in computer vision, and it aims to distinguish different categories of images. 
Some works \cite{Resnet,VGG} employ convolutional Neural Networks (CNNs) for image classification, due to the convolutional layer' capability to capture local features within images. For instance, AlexNet \cite{Alexnet}, consisting of five convolutional layers and three fully connected layers, achieves great image classification performance. VGG \cite{VGG} and ResNet \cite{Resnet} respectively propose enhancements by increasing the depth of the original network and integrating skip connections to further enhance the model's classification capabilities.
However, CNNs have limitations in understanding global information and lack robustness when capturing global and long-distance dependencies \cite{ViT}. Vision Transformer (ViT) \cite{ViT} leverages multi-head self-attention (MSA) \cite{Transformer} to capture context information of each patch, which enhances the model's capability to capture global dependencies. Moreover, Swin Transformer \cite{Swin} adopts a windowed self-attention mechanism and hierarchical structural design, which not only retains the global modeling capabilities of the MSA but also enhances the extraction of local features. Furthermore, MLP-Mixer \cite{MLPMixer} proposed a pure MLP-based architecture to capture different contextual relationships and enhance visual representation. In addition, VMamba \cite{VMamba} improves visual classification tasks by integrating a novel Sequence State Space (S4) model with a Selection mechanism and Scan computation, termed Mamba.

\subsection{Insect Pest Classification}
For the insect pest classification task, it can help people better understand the population dynamics, and potential damage of pests, to formulate effective pest management strategies, which is very important for agriculture economy, and environmental science.  
However, compared to general images, the feature differences in the insect pest domain may be very subtle, and the background is more complex, which places higher requirements on the classification model and requires more accurate extraction of effective features \cite{doan2023large,Hieu2021}.
For this challenge, some works \cite{cheng2017,liu2016localization,wang2017crop,ThenmozhiR19,RenLW19} improve CNN-based models to capture pest features under a complex background. 
In addition, Faster-PestNet \cite{FasterPestNet} used MobileNet \cite{MobileNet} to extract sample attributes, and redesigned an improved Faster-RCNN method to recognize the crop pests.
\citet{Hieu2021} propose a CNN-based model with an attention mechanism to further focus on insects in the image; \citet{an2023insect} proposes a feature fusion network that synthesizes representations from different backbone models to enhance insect image classification; \citet{anwar2023exploring} employ deep ensemble models method \cite{HuZW0L23} to improve accuracy and robustness in insect and pest detection from images.
Moreover, \citet{PengW22} investigated ViT architecture in the insect domain and aggregated CNNs and self-attention models to further improve capability for insect pest classification.

\section{Preliminaries}\label{sec:preliminaries}
\subsection{Convolutional Neural Networks}
Convolutional Neural Networks (CNNs \cite{CNN}) are widely applied to computer vision owing to their strong capability for image feature extraction. 
It consists of a set of fixed-size learnable parameters known as filters and continuously performs convolutional computations with a sliding window across the input images. 
Specifically, given visual features $\mV \in \R^{H \times W \times C}$, where $H$, $W$, and $C$ are the height, width, and number of channels, we can use convolution kernels $w$ with a size of $F_w$, $F_h$, $C_{in}$ to calculate the pixel value of each channel for the visual features, i.e.,
\begin{align}
\mV_{out}[i, j, k] = \sum_{l=0}^{C_{in}-1} \left( \sum_{m=0}^{F_h-1} \sum_{n=0}^{F_w-1} \mV[i \times S + m, j \times S + n, l] \times w[m, n, l, k] \right) + b[k]
\end{align}
where $\mV_{out}$ is output feature map, $(i,j,k)$ is the index, $S$ is stride, and $b[k]$ is the bias of channel $k$. Through the cascading structure, CNN can gradually learn from low-level to high-level feature representations from the original data, and finally achieve effective classification.

\subsection{Multi-Head Self-Attention}
Multi-Head Self-Attention (MSA) is proposed by \citet{Transformer} and is widely used for many natural language processing tasks \cite{zhou2023thread,han2024chainofinteraction,liu2024news}. Unlike convolutional neural networks, MSA allows the model to weigh the importance of different input tokens when generating output representations, enabling the model to capture global dependencies and contextual information within the sequence effectively. Recently, Transformer-like architectures have also demonstrated powerful modeling capabilities in computer vision \cite{ViT}. Specifically, given visual features $\mV \in \R^{H \times W \times C}$, the multi-head self-attention modeling of the visual features can be defined as: 
\begin{align}
{\Attn}_t^h = \softmax(\frac{\mQ_t^h\cdot(\mK_t^h)^T}{\sqrt{d_{\mK_t^h}}}), \text{Where}~ \mQ_t^h = \mW_Q^h\cdot\mV, \mK_t^h = \mW_k^h\cdot\mV,
\end{align}
where $\mW_Q^h \in \R^{D \times d}$ and $\mW_k^h \in \R^{D \times d}$ refer to linear projections, which project the D-dimensional input vector into query $\mQ_t^h\in \R^{N \times d}$ and key $\mK_t^h\in \R^{N \times d}$, respectively. Each Attention matrix ${\Attn}_t^h$ is used to multiply value to obtain updated representation that fused global information, i.e.,
\begin{align}
\mV := {\Attn}_t^h \cdot \mV_t^h, \text{Where}~ \mV_t^h=\mW_t^h \cdot \mV
\end{align}
In vision tasks, MSA needs to be pre-trained on large-scale datasets to make up for its lack of inductive bias in CNN, such as translation invariance and locality.

\subsection{Multi-Layer Perceptron}
Multi-layer perceptron(MLP) is a commonly used neural network layer for many tasks \cite{MLP,MLPMixer}. An MLP mainly contains N linear layers, each layer has learnable parameters of both weight and bias as well as activation functions. The activation function is used to map the non-linear relationship between input and output. Specifically, given visual features $\mV \in \R^{H \times W \times C}$, MLP, only associated with channels, maps each channel to a $D$-dimensional hidden vector $\vh_i$. i.e.,
\begin{align}
\vh_i = \Act(\sum\limits_{j}\mW_{ij} * c_j + b_i), \mW \in \R^{C\times D}, b\in \R^{D},
\end{align}
where $\vh_i$ is the $i$-th dimension of $\mH$, obtained by weighting  $C$ channels and learnable parameters  in the $i$-th column of weight matrix $\mW$. $\Act$ is an activation function, that adjusts the output through non-linear transformation.

\subsection{State Space Models} 
State Space Models (SSMs) \cite{Mamba,VMamba} introduce a novel Cross-Scan Module (CSM) for improved directional sensitivity and computational efficiency. SSMs are pivotal in modeling the dynamics of visual systems through equations that describe temporal evolution and observation generation. The observation function is as follows:
\begin{align}
\vx_{t+1} = \mA \cdot \vx_t + \mB \cdot \vu_t + \vw_t,
\end{align}
where $\vx_t$ denotes the system state at time $t$, $\vu_t$ represents control inputs, and $\vw_t$ is the process noise, indicating uncertainties in state transitions.
Moreover, the observation function can be defined as:
\begin{align}
\vy_t = \mC \cdot \vx_t + \mD \cdot \vu_t + \vv_t,
\end{align}
with $\vy_t$ as the observation at time $t$, and $\vv_t$ as the observation noise, highlighting discrepancies between modeled and actual observations. Matrices $\mA, \mB, \mC, \mD$ define the dynamics, linking state transitions to observations. Furthermore, the Cross-Scan Module (CSM) further addresses directional sensitivity by structuring visual features into ordered patch sequences through:
\begin{align}
\CSM(\mV) = \Order(\Traverse(\mV)),
\end{align}
where $\mV$ is a visual feature input. This process allows for effective spatial information handling, improving the model's dynamic processing capabilities.

\section{InsectMamba}
This section elaborates on the architecture of our InsectMamba Model, a novel vision model for insect pest classification. The backbone of our model is the Mix-SSM Block, designed to integrate features from various visual encoding strategies. Finally, we introduce our proposed Selective Module, which can integrate adaptively representations derived from different visual encoding strategies.

\subsection{Overall Architecture}
\begin{figure}
    \centering
    \includegraphics[width=1\linewidth]{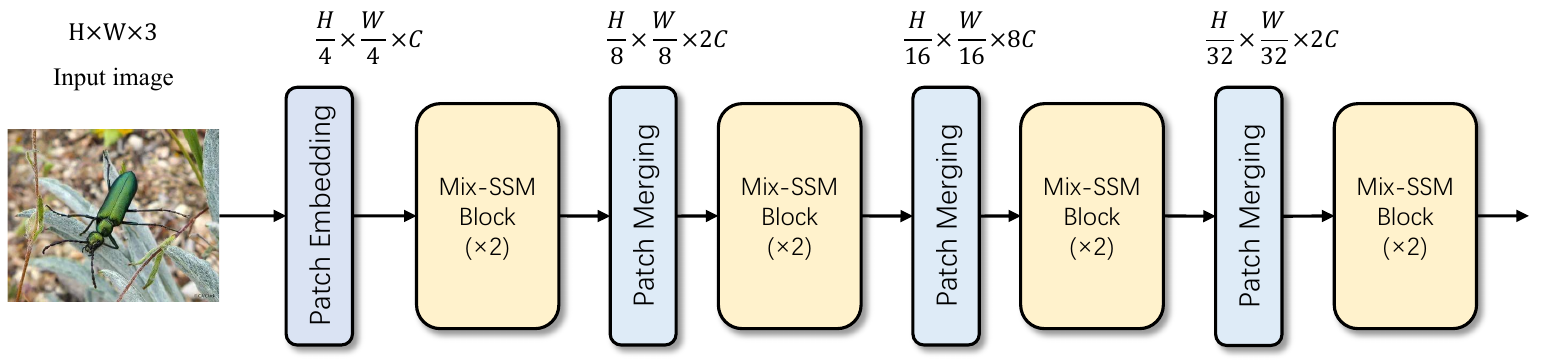}
    \caption{The overall architecture of our InsectMamba Model.}
    \label{fig:model}
\end{figure}
As shown in Figure~\ref{fig:model}, given an image $\mI \in \R^{H \times W \times 3}$, the image is initially segmented into multiple non-overlapping $4 \times 4$ patches. Subsequently, a patch embedding layer \cite{ViT} is utilized to transform these patches into a lower-dimensional latent space, resulting in dimensions $\frac{H}{4} \times \frac{W}{4} \times C$, where $C$ denotes the number of channel in the latent space, i.e.,
\begin{align}
\mV = \patch(\mI), \mV \in \R^{\frac{H}{4} \times \frac{W}{4} \times C} \label{equ:patch}.
\end{align}
Subsequently, we pass the features $\mV$ into Mix-SSM Blocks for feature extraction, and dimensionality reduction is achieved through a Patch Merging operation \cite{Swin}, i.e.,
\begin{align}
\mV := \merging(\mix(\mV)) \label{equ:mix}.
\end{align}
After several iterations of Mix-SSM Blocks and Patch Merging operations shown in Figure~\ref{fig:model}, the final visual representation of the image, $\vv \in \R^{L}$, is derived. lastly, $\vv$ is passed through a linear layer $\text{Linear}$ to transform its dimensions to the number of classes, i.e.,
\begin{align}
\vh = \linear(\vv), \label{equ:clf} \notag \\
\vp = \softmax(\vh)
\end{align}
where $\softmax$ converts the hidden features $\vh$ into a probability distribution over each class $\vp$.

\subsection{Mix-SSM Block}
\begin{figure}
    \centering
    \includegraphics[width=1\linewidth]{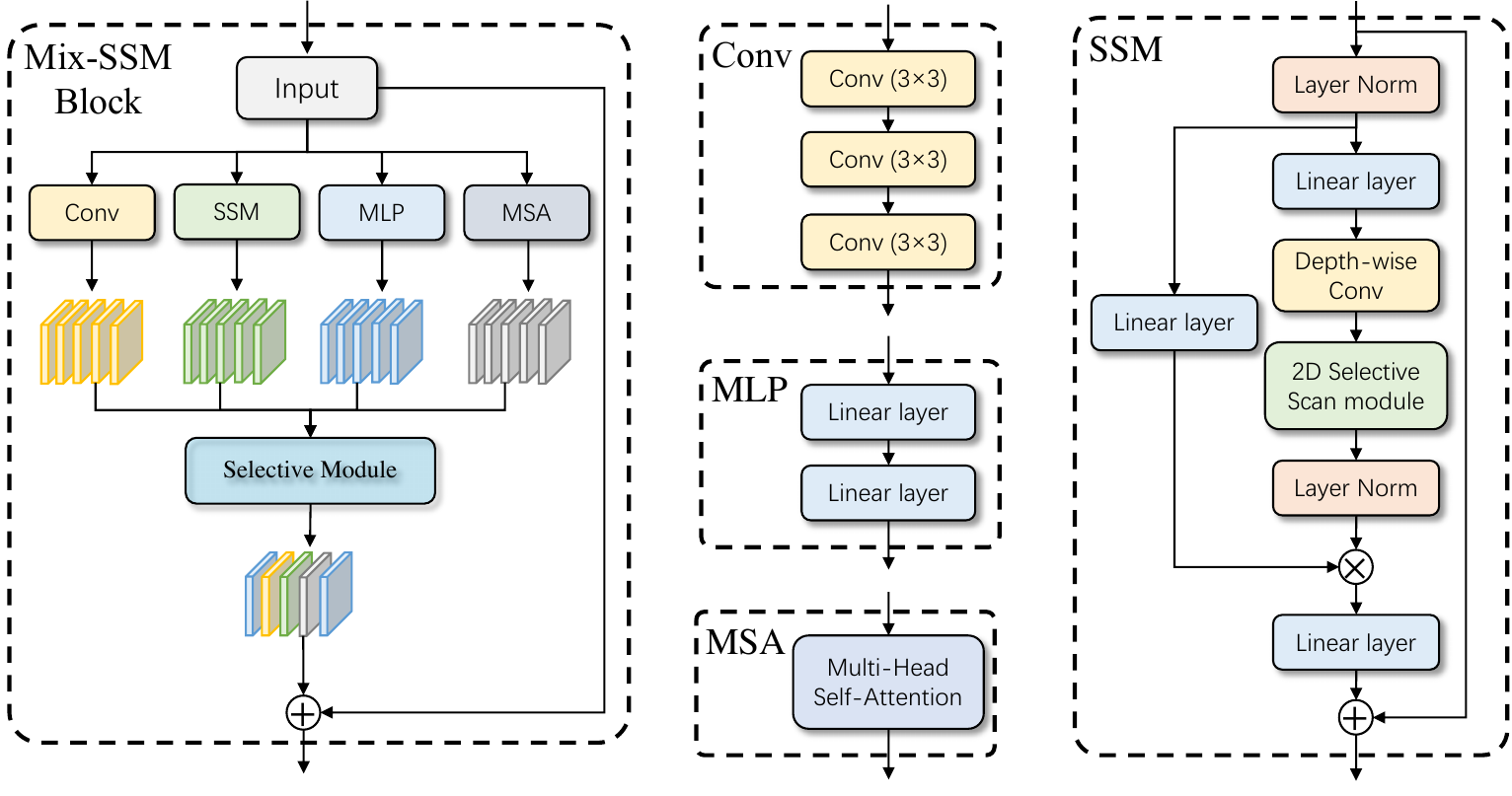}
    \caption{Details of our Mix-SSM Block.}
    \label{fig:module}
\end{figure}
The Mix-SSM Block is composed of several key components: a Selective Scan Module (SSM), convolutional layers (Conv), a Multi-Layer Perceptron (MLP), a Multi-Head Self-Attention mechanism (MSA), and a Selective Module. The details of the different kinds of visual encoding strategies, i.e., SSM, Conv, MLP, and MSA, can be found in Section \ref{sec:preliminaries}.

As shown in Figure~\ref{fig:module}, 
Given features $\mV$ from Equation~\ref{equ:patch}, we pass it into Mix-SSM Blocks. 
The features $\mV$ are respectively encoded with different visual encoding strategies, and we obtain $\mF_{\SSM}$, $\mF_{\Conv}$, $\mF_{\MLP}$, and $\mF_{\MSA}$, i.e., 
\begin{align}
\mF_{\SSM} = \SSM(\mV), \notag \\
\mF_{\Conv} = \Conv(\mV), \notag \\
\mF_{\MLP} = \MLP(\mV), \notag \\
\mF_{\MSA} = \MSA(\mV).
\end{align}
where $\mF_m, m \in \{\SSM, \Conv, \MLP, \MSA\}$, is encoded features in same dimension. $\SSM$ aims to adaptively aggregate spatial information with long-distance dependencies based on the input features, $\Conv$ plays its role in extracting local visual features, $\MLP$ processes the channel-wise information, and $\MSA$ captures global dependencies for visual features.
Subsequently, the features $\mF_{\SSM}$, $\mF_{\Conv}$, $\mF_{\MLP}$, and $\mF_{\MSA}$ are passed through the Selective Module to adaptively aggregate the features. 

\subsection{Selective Module}
To integrate features from different encoding strategies and leverage their respective advantages, we introduce the Selective Module, which enables the model to adaptively adjust visual features across different encoding strategies. Specifically, we first integrate features $\mF_{\SSM}$, $\mF_{\Conv}$, $\mF_{\MLP}$, and $\mF_{\MSA}$ from different encoding strategies as follows:
\begin{align}
\mF = \mF_{\SSM} + \mF_{\Conv} + \mF_{\MLP} + \mF_{\MSA}.
\end{align}
Then, we aggregate the information across each channel by employing global average pooling to obtain embedded global features $\mF \in \R^{\bar{C}}$. $\bar{C}$, $\bar{W}$, and $\bar{H}$ denote the number of feature channels, width, and height entering the Selective Module within Mix-SSM, respectively.
In particular, the $c$-th element of $\vg$ is computed by spatially downsampling $\mF_c$ over dimensions $\bar{H} \times \bar{W}$:
\begin{align}
\vg_c = \globalpool(\mF_c) = \frac{1}{\bar{H} \times \bar{W}} \sum_{i=1}^{\bar{H}} \sum_{j=1}^{\bar{W}} \mF_c(i, j). \label{equ:pool}
\end{align}
To enable the model to infer the weight of different encoding strategies for various channels, we further encode $\vg$ using an MLP to obtain hidden features $\vh$:
\begin{align}
\vh = {\MLP}_h(\vg), \vh \in \R^{\bar{C} \times n}
\end{align}
where $n$ represents the number of visual encoding strategies. Cross-channel soft attention is applied to adaptively select information across different spatial scales. Specifically, a softmax operation is applied to the channel dimensions of hidden features $\vh$:
\begin{align}
\vp = \softmax(\vh), \vp \in \R^{\bar{C} \times n}
\end{align}
where $\vp$ signifies the weight of different encoding strategies across each channel. Weighting the features $\mF_{\SSM}$, $\mF_{\Conv}$, $\mF_{\MLP}$, and $\mF_{\MSA}$ based on $\{\vp_{\SSM}, \vp_{\Conv}, \vp_{\MLP}, \vp_{\MSA}\} \in \vp$ to obtain $\mV$:
\begin{align}
\mV := \vp_{\SSM} \cdot \mF_{\SSM} + \vp_{\Conv} \cdot \mF_{\Conv} + \vp_{\MLP} \cdot \mF_{\MLP} + \vp_{\MSA} \cdot \mF_{\MSA}.
\end{align}

\section{Experiment}
In experiments, we evaluate the performance of our InsectMamba model on five insect pest classification datasets. We compare the performance of our model with several state-of-the-art models. We also conduct an ablation study to investigate the effectiveness of different components in our model.

\subsection{Dataset and Metrics}
\begin{wraptable}{r}{0.5\textwidth}\small
\vspace{-0.5cm}
\centering
\caption{Statistics of five insect pest classification datasets.}
\label{tab:dataset}
\setlength{\tabcolsep}{2.3pt}
\begin{tabular}{lccc}
\toprule
\textbf{Dataset} & \textbf{Category} & \textbf{Train} & \textbf{Test} \\
\midrule
Farm Insects & 15 & 160 & 1,368 \\
Agricultural Pests & 12 & 240 & 5,254 \\
Insect Recognition & 24 & 768 & 612 \\
Forestry Pest Identification & 31 & 599 & 6,564 \\
IP102 & 102 & 1,909 & 65,805 \\
\bottomrule
\end{tabular}
\vspace{-0.3cm}
\end{wraptable}

To more effectively and comprehensively evaluate existing visual models, we curated and re-split five insect pest classification datasets to provide a challenging evaluation. The datasets employed in our experiments are Farm Insects \footnote{\url{https://www.kaggle.com/datasets/tarundalal/dangerous-insects-dataset}}, Agricultural Pests \footnote{\url{https://www.kaggle.com/datasets/gauravduttakiit/agricultural-pests-dataset}}, Insect Recognition \cite{InsectRecognition}, Forestry Pest Identification \cite{ForestryPestIdentification}, and IP102 \cite{IP102}, with details provided in Table~\ref{tab:dataset}. We reduce the number of samples in the training set to compare the encoding capabilities of different visual models for visual features. In addition, we leverage accuracy (ACC), precision (Prec), recall (Rec), and the F1 score as evaluation metrics to evaluate the performance of the models comprehensively.

\subsection{Implementation Details}
For our model training, the batch size is configured to 32, and the learning rate was set at $5 \times 10^{-5}$. We conduct training on 10 epochs using the Adam optimizer \cite{Adam}. The dimensions of the input images were fixed at $224 \times 224 \times 3$ pixels. For comparative analysis, we finetune various models on five datasets, i.e., ResNet18 \cite{Resnet}, ResNet50 \cite{Resnet}, ResNet101 \cite{Resnet}, ResNet152 \cite{Resnet}, DeiT-S \cite{DeiT}, DeiT-B \cite{DeiT}, Swin-T \cite{Swin}, Swin-S \cite{Swin}, Swin-B \cite{Swin}, Vmamba-T \cite{VMamba}, Vmamba-S \cite{VMamba}, and Vmamba-B \cite{VMamba}. ``T'', ``S'', and ``B'' denote the ``Tiny'', ``Small'', and ``Base'' model size of corresponding models, respectively. To initialize parts of our model's parameters, we utilized pre-trained parameters from Vmamba-B.

\subsection{Main Results}
\begin{table}[t]\small
\centering
\begin{minipage}{.48\linewidth}
\centering
\caption{Model performance comparison on Farm Insects Dataset.}
\label{tab:main_farm_insects}
\begin{tabular}{lcccc}
\toprule
Method & ACC & Prec & Rec & F1 \\
\midrule
ResNet18 & 0.43 & 0.51 & 0.43 & 0.42 \\
ResNet50 & 0.50 & 0.54 & 0.50 & 0.47 \\
ResNet101 & 0.52 & 0.55 & 0.52 & 0.49 \\
ResNet152 & 0.53 & 0.59 & 0.53 & 0.50 \\
DeiT-S & 0.49 & 0.50 & 0.50 & 0.47 \\
DeiT-B & 0.55 & 0.59 & 0.55 & 0.54 \\
Swin-T & 0.54 & 0.53 & 0.54 & 0.52 \\
Swin-S & 0.56 & 0.60 & 0.56 & 0.55 \\
Swin-B & 0.62 & \bf 0.68 & 0.63 & 0.63 \\
Vmamba-T & 0.56 & 0.58 & 0.56 & 0.55 \\
Vmamba-S & 0.53 & 0.56 & 0.53 & 0.51 \\
Vmamba-B & 0.52 & 0.56 & 0.52 & 0.48 \\
InsectMamba & \bf 0.66 & 0.67 & \bf 0.66 & \bf 0.65 \\
\bottomrule
\end{tabular}
\end{minipage}
\hfill
\begin{minipage}{.48\linewidth}
\centering
\caption{Model Performance Comparison on Agricultural Pests Dataset.}
\label{table:main_Agricultural_Pests}
\begin{tabular}{lcccc}
\toprule
Method & ACC & Prec & Rec & F1 \\
\midrule
ResNet18 & 0.69 & 0.71 & 0.67 & 0.64 \\
ResNet50 & 0.68 & 0.77 & 0.67 & 0.65 \\
ResNet101 & 0.75 & 0.76 & 0.74 & 0.73 \\
ResNet152 & 0.78 & 0.80 & 0.76 & 0.76 \\
DeiT-S & 0.83 & 0.82 & 0.82 & 0.82 \\
DeiT-B & 0.86 & 0.86 & 0.85 & 0.85 \\
Swin-T & 0.80 & 0.82 & 0.80 & 0.79 \\
Swin-S & 0.74 & 0.79 & 0.73 & 0.74 \\
Swin-B & 0.83 & 0.86 & 0.83 & 0.82 \\
Vmamba-T & 0.78 & 0.81 & 0.77 & 0.77 \\
Vmamba-S & 0.83 & 0.84 & 0.82 & 0.80 \\
Vmamba-B & 0.89 & 0.90 & 0.89 & 0.89 \\
InsectMamba & \bf 0.91 & \bf 0.91 & \bf 0.90 & \bf 0.91 \\
\bottomrule
\end{tabular}
\end{minipage}
\end{table}

\begin{table}[t]\small
\centering
\begin{minipage}{.48\linewidth}
\centering
\caption{Model performance comparison on Insect Recognition Dataset.}
\label{table:main_insect_recognition}
\begin{tabular}{lcccc}
\toprule
Method & ACC & Prec & Rec & F1 \\
\midrule
ResNet18 & 0.66 & 0.69 & 0.66 & 0.64 \\
ResNet50 & 0.57 & 0.73 & 0.57 & 0.56 \\
ResNet101 & 0.57 & 0.66 & 0.57 & 0.55 \\
ResNet152 & 0.52 & 0.67 & 0.52 & 0.52 \\
DeiT-S & 0.73 & 0.76 & 0.74 & 0.73 \\
DeiT-B & 0.76 & 0.82 & 0.76 & 0.76 \\
Swin-T & 0.70 & 0.81 & 0.70 & 0.69 \\
Swin-S & 0.75 & 0.80 & 0.76 & 0.76 \\
Swin-B & 0.81 & 0.86 & 0.82 & 0.82 \\
Vmamba-T & 0.79 & 0.85 & 0.79 & 0.79 \\
Vmamba-S & 0.81 & 0.84 & 0.80 & 0.79 \\
Vmamba-B & 0.83 & 0.87 & 0.84 & 0.84 \\
InsectMamba & \bf 0.86 & \bf 0.88 & \bf 0.86 & \bf 0.86 \\
\bottomrule
\end{tabular}
\end{minipage}
\hfill
\begin{minipage}{.48\linewidth}
\centering
\caption{Model performance comparison on Forestry Pest Identification Dataset.}
\label{tab:main_forestry_pest_identification}
\begin{tabular}{lcccc}
\toprule
Method & ACC & Prec & Rec & F1 \\
\midrule
ResNet18 & 0.80 & 0.83 & 0.80 & 0.80 \\
ResNet50 & 0.80 & 0.84 & 0.80 & 0.80 \\
ResNet101 & 0.79 & 0.85 & 0.79 & 0.79 \\
ResNet152 & 0.77 & 0.82 & 0.77 & 0.76 \\
DeiT-S & 0.87 & 0.88 & 0.87 & 0.87 \\
DeiT-B & 0.90 & 0.91 & 0.90 & 0.90 \\
Swin-T & 0.83 & 0.88 & 0.84 & 0.84 \\
Swin-S & 0.85 & 0.88 & 0.85 & 0.85 \\
Swin-B & 0.86 & 0.89 & 0.87 & 0.86 \\
Vmamba-T & 0.90 & 0.91 & 0.90 & 0.90 \\
Vmamba-S & 0.91 & 0.92 & 0.92 & 0.91 \\
Vmamba-B & 0.92 & 0.93 & 0.93 & 0.93 \\
InsectMamba & \bf 0.94 & \bf 0.94 & \bf 0.94 & \bf 0.94 \\
\bottomrule
\end{tabular}
\end{minipage}
\end{table}

\begin{wraptable}{r}{0.46\textwidth}\small
\vspace{-1.55cm}
\centering
\caption{Model performance comparison on IP102 Dataset.}
\label{tab:main_ip102}
\begin{tabular}{@{}lcccc@{}}
\toprule
Method      & ACC & Prec & Rec & F1   \\ \midrule
ResNet18    & 0.27     & 0.27      & 0.25   & 0.21 \\
ResNet50    & 0.24     & 0.26      & 0.22   & 0.18 \\
ResNet101   & 0.18     & 0.28      & 0.19   & 0.16 \\
ResNet152   & 0.25     & 0.23      & 0.19   & 0.16 \\
DeiT-S      & 0.22     & 0.24      & 0.21   & 0.17 \\
DeiT-B      & 0.28     & 0.27      & 0.25   & 0.20 \\
Swin-T      & 0.29     & 0.25      & 0.27   & 0.22 \\
Swin-S      & 0.30     & 0.25      & 0.27   & 0.21 \\
Swin-B      & 0.39     & 0.36      & 0.37   & 0.32 \\
Vmamba-T    & 0.28     & 0.29      & 0.29   & 0.23 \\
Vmamba-S    & 0.35     & 0.31      & 0.34   & 0.27 \\
Vmamba-B    & 0.32     & 0.36      & 0.33   & 0.28 \\
InsectMamba & \bf 0.43     & \bf 0.38      & \bf 0.42   & \bf 0.37 \\ \bottomrule
\end{tabular}
\vspace{-0.6cm}
\end{wraptable}
The experimental results, as shown in Tables~\ref{tab:main_farm_insects}, \ref{table:main_Agricultural_Pests}, \ref{table:main_insect_recognition}, \ref{tab:main_forestry_pest_identification}, and \ref{tab:main_ip102}, demonstrate the superior performance of the InsectMamba model across multiple insect classification tasks. InsectMamba consistently outperforms the established benchmarks, including various configurations of ResNet, DeiT, Swin Transformer, and Vmamba, across all evaluation metrics: Accuracy (ACC), Precision (Prec), Recall (Rec), and F1 Score (F1). On the Farm Insects Dataset, InsectMamba achieves an ACC of 0.66, surpassing the next best model, Swin-B, by 4\%. Similarly, significant improvements are observed on the Agricultural Pests Dataset, where InsectMamba reaches an ACC of 0.91, outperforming the strong Vmamba-B baseline by 2\%. These results are consistent across the Insect Recognition and Forestry Pest Identification Datasets, which shows InsectMamba's strong capability to extract features for images. The results on the IP102 Dataset further verify InsectMamba's robustness, achieving an ACC of 0.43, which is a leap over the previous best of 0.39 by Swin-B. These results demonstrate that the Mix-SSM Block can integrate multiple visual encoding strategies to ensure comprehensive feature capture from the input images. The Selective Module further enhances the model's capability by adaptively weighting the contribution of different encoding strategies.

\subsection{Ablation Study}
\begin{table}[t]
\centering
\caption{Ablation study of InsectMamba.}
\label{tab:ablation}
\begin{tabular}{lcccccc}
\toprule
\multirow{2}{*}{Method} & \multicolumn{2}{c}{Farm Insects} & \multicolumn{2}{c}{Insect Recognition} & \multicolumn{2}{c}{IP102} \\ \cmidrule(lr){2-3} \cmidrule(lr){4-5} \cmidrule(lr){6-7}
 & Accuracy & F1 & Accuracy & F1 & Accuracy & F1 \\ \midrule
InsectMamba & \bf 0.66 & \bf 0.65 & \bf 0.86 & \bf 0.86 & \bf 0.43 & \bf 0.37 \\ \midrule
w/o CNN & 0.60 & 0.58 & 0.84 & 0.85 & 0.38 & 0.33 \\
w/o MSA & 0.62 & 0.60 & 0.85 & 0.86 & 0.40 & 0.34 \\
w/o MLP & 0.63 & 0.60 & 0.85 & 0.85 & 0.41 & 0.34 \\
w/o CNN, MSA & 0.54 & 0.50 & 0.83 & 0.84 & 0.34 & 0.29 \\
w/o CNN, MLP & 0.55 & 0.53 & 0.84 & 0.84 & 0.34 & 0.30 \\
w/o MSA, MLP & 0.57 & 0.55 & 0.84 & 0.84 & 0.35 & 0.31 \\
w/o CNN, MSA, MLP & 0.52 & 0.48 & 0.83 & 0.84 & 0.32 & 0.28 \\ \bottomrule
\end{tabular}
\end{table}
The ablation study results are shown in Table~\ref{tab:ablation} systematically evaluates the contribution of each component within the InsectMamba model, namely, Convolutional Neural Networks (CNN), Multi-Layer Perceptron (MLP), and Multi-Head Self-Attention (MSA), across three datasets: Farm Insects, Insect Recognition, and IP102. The results highlight the significant role each component plays in achieving high accuracy and F1 scores. The complete InsectMamba model achieves the best performance across all datasets, which underscores the synergistic effect of combining CNN, MLP, and MSA for feature extraction and representation learning. Removing any single component (CNN, MSA, or MLP) leads to a decrease in both accuracy and F1 scores across all datasets, indicating that each component contributes unique and valuable information for classification. The most significant performance degradation is observed when multiple components are removed simultaneously, particularly when CNN, MSA, and MLP are all excluded. This configuration results in the lowest accuracy and F1 scores, demonstrating that the integration of multiple visual encoding strategies is crucial for capturing the comprehensive visual characteristics of insects.

\subsection{Analysis}
\paragraph{Impact of Feature Aggregation Methods.}
\begin{figure}[t]
    \centering
    \begin{subfigure}{0.48\linewidth}
        \centering
        \includegraphics[width=\linewidth]{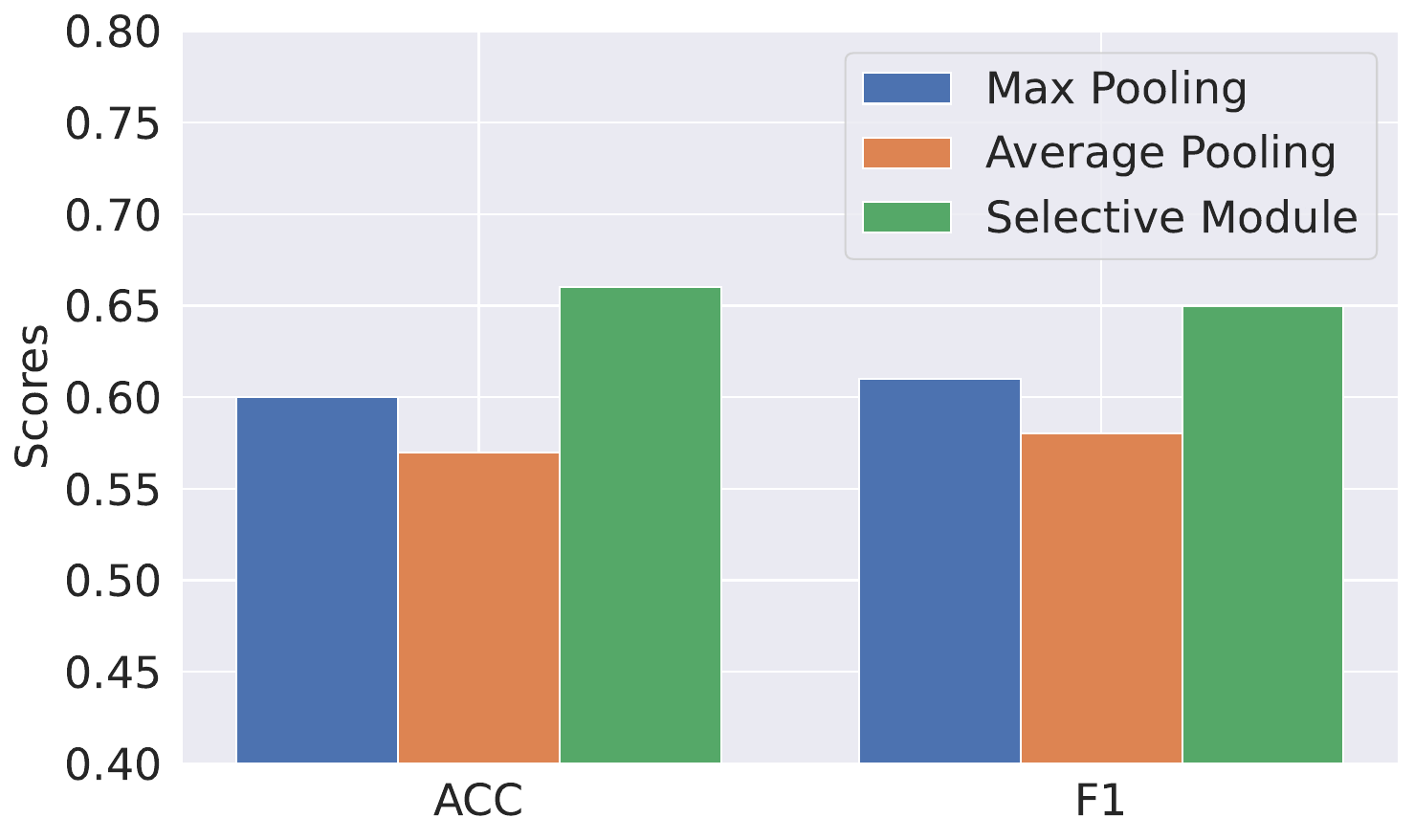}
        \caption{Farm Insects}
    \end{subfigure}
    \hfill
    \begin{subfigure}{0.48\linewidth}
        \centering
        \includegraphics[width=\linewidth]{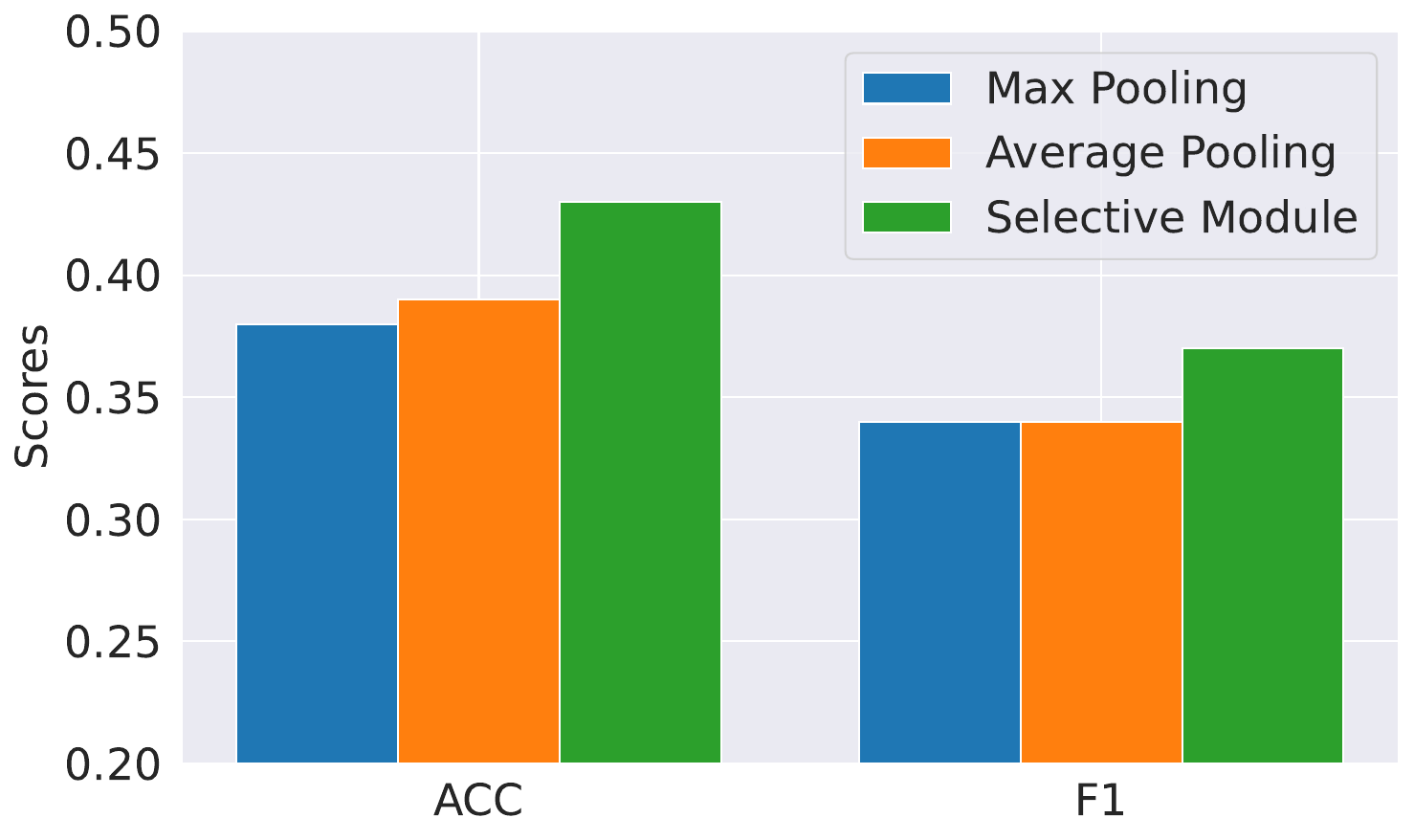}
        \caption{IP102}
    \end{subfigure}
    \caption{Comparison of feature aggregation methods for different encoding blocks.}
    \label{fig:feature}
\end{figure}
To investigate the effectiveness of different feature aggregation methods within the InsectMamba model, we evaluate by comparing the Selective Module against Max Pooling and Average Pooling methods. As depicted in Figure~\ref{fig:feature}, the Selective Module consistently outperforms Max Pooling and Average Pooling in terms of Accuracy (ACC) and F1 Score across two distinct datasets: Farm Insects and IP102. For the Farm Insects dataset, the Selective Module achieves the highest ACC and F1 Score, indicating its superior capability in capturing and integrating salient features for insect pest classification. Specifically, the ACC and F1 improvement over Max Pooling is pronounced, underlining the Selective Module's effectiveness in handling more nuanced classification tasks within a diverse set of insect species. On the IP102 dataset, the Selective Module still maintains an advantage. Moreover, the variance in performance across the two datasets also highlights the adaptive nature of the Selective Module. It demonstrates that the Selective Module can dynamically adjust the integration of visual features from different visual encoding strategies according to the dataset's complexity and diversity.

\paragraph{Impact of kernel size in the selective module.}
\begin{figure}[t]
    \centering
    \begin{subfigure}{0.48\linewidth}
        \centering
        \includegraphics[width=\linewidth]{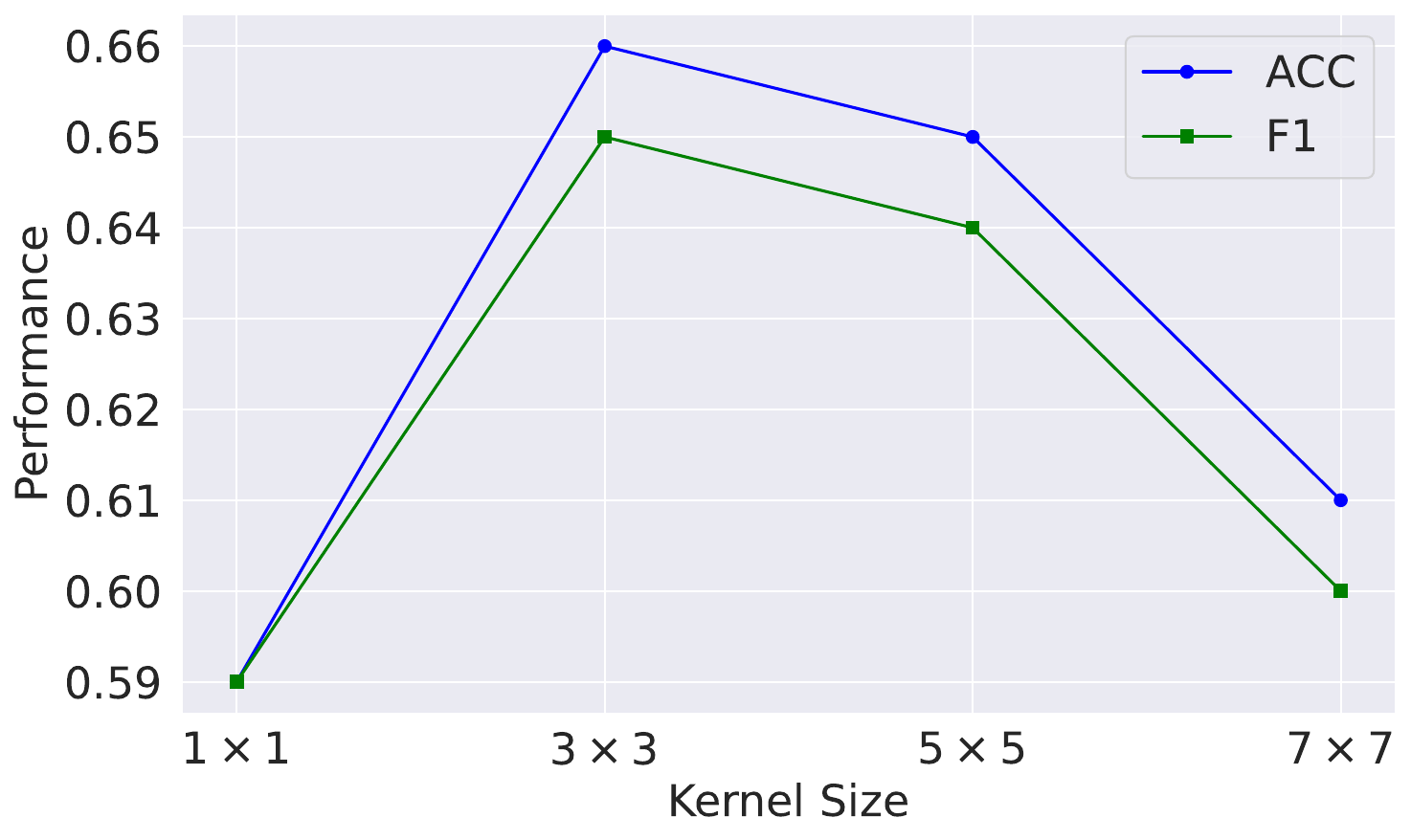}
        \caption{Farm Insects}
    \end{subfigure}
    \hfill
    \begin{subfigure}{0.48\linewidth}
        \centering
        \includegraphics[width=\linewidth]{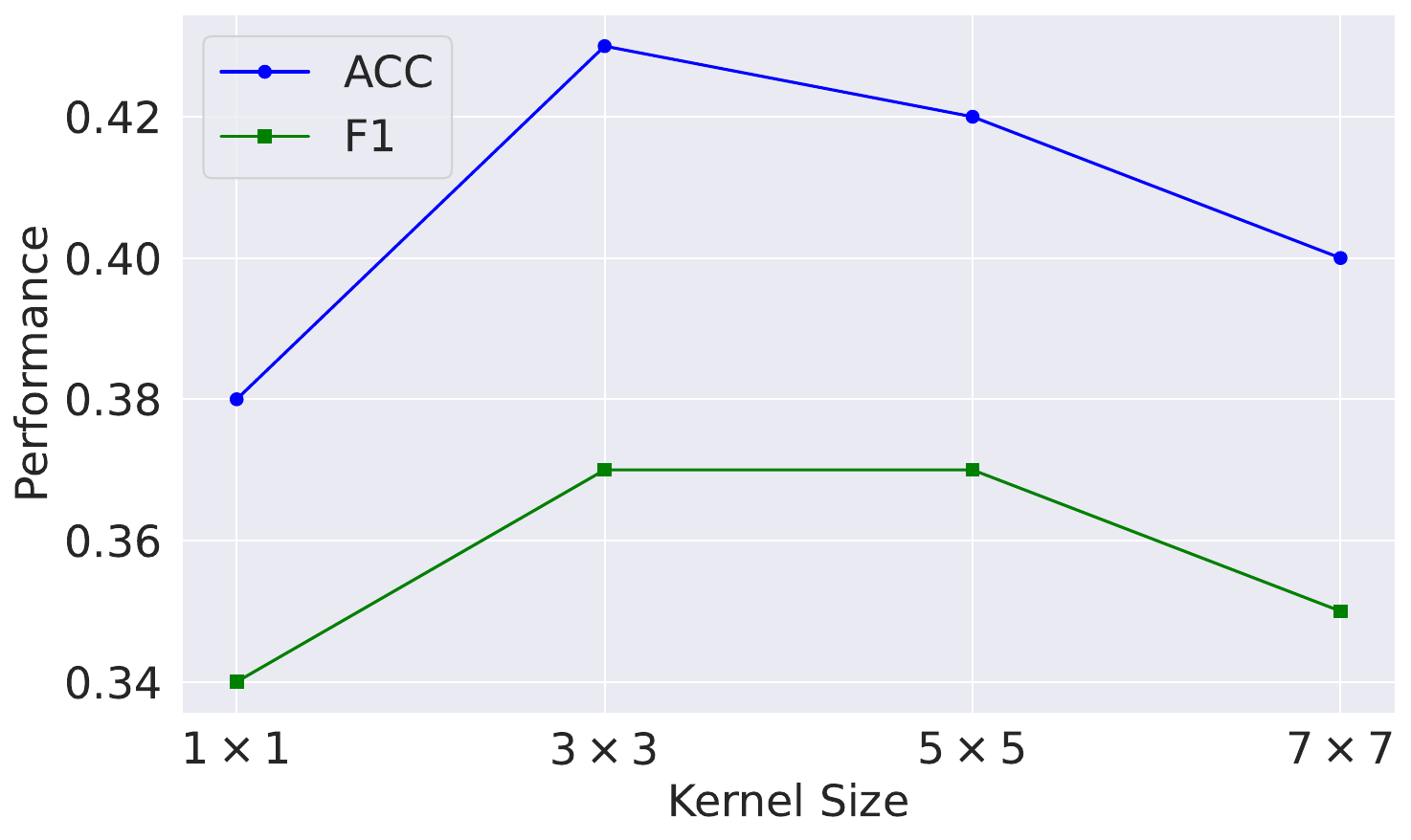}
        \caption{IP102}
    \end{subfigure}
    \caption{Investigation of different kernel sizes in the selective module.}
    \label{fig:kernel}
\end{figure}
In the process of optimizing our InsectMamba model, we investigated the impact of different kernel sizes within the Selective Module on the classification performance. As shown in Figure~\ref{fig:kernel}, the Selective Module was evaluated with kernel sizes of $1 \times 1$, $3 \times 3$, $5 \times 5$, and $7 \times 7$. For the Farm Insects dataset, shown in Figure~\ref{fig:kernel}(a), we observe that both Accuracy (ACC) and F1 Score (F1) metrics peak at a kernel size of $3 \times 3$. The performance declines when the kernel size is increased to $5 \times 5$ and drops significantly at $7 \times 7$. It demonstrates that smaller kernel sizes are more effective at capturing the relevant visual features for insect pest classification. Moreover, the IP102 dataset, shown in Figure~\ref{fig:kernel}(b), shows a consistent trend. It shows the importance of the kernel sizes in the Selective Module in adaptively integrating different visual encoding strategies.

\paragraph{Impact of pooling methods in the selective module.}
\begin{figure}[t]
    \centering
    \begin{subfigure}{0.48\linewidth}
        \centering
        \includegraphics[width=\linewidth]{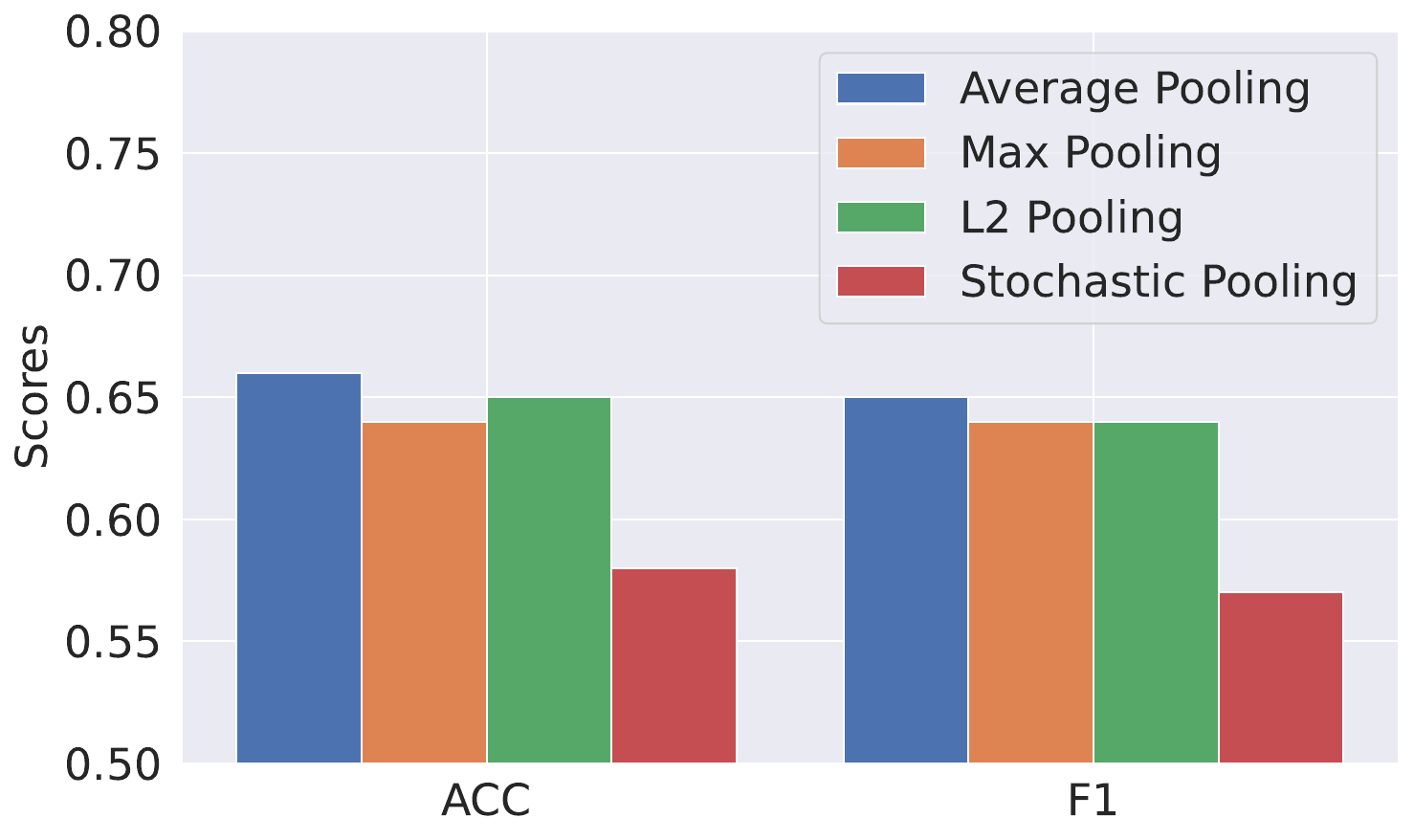}
        \caption{Farm Insects}
    \end{subfigure}
    \hfill
    \begin{subfigure}{0.48\linewidth}
        \centering
        \includegraphics[width=\linewidth]{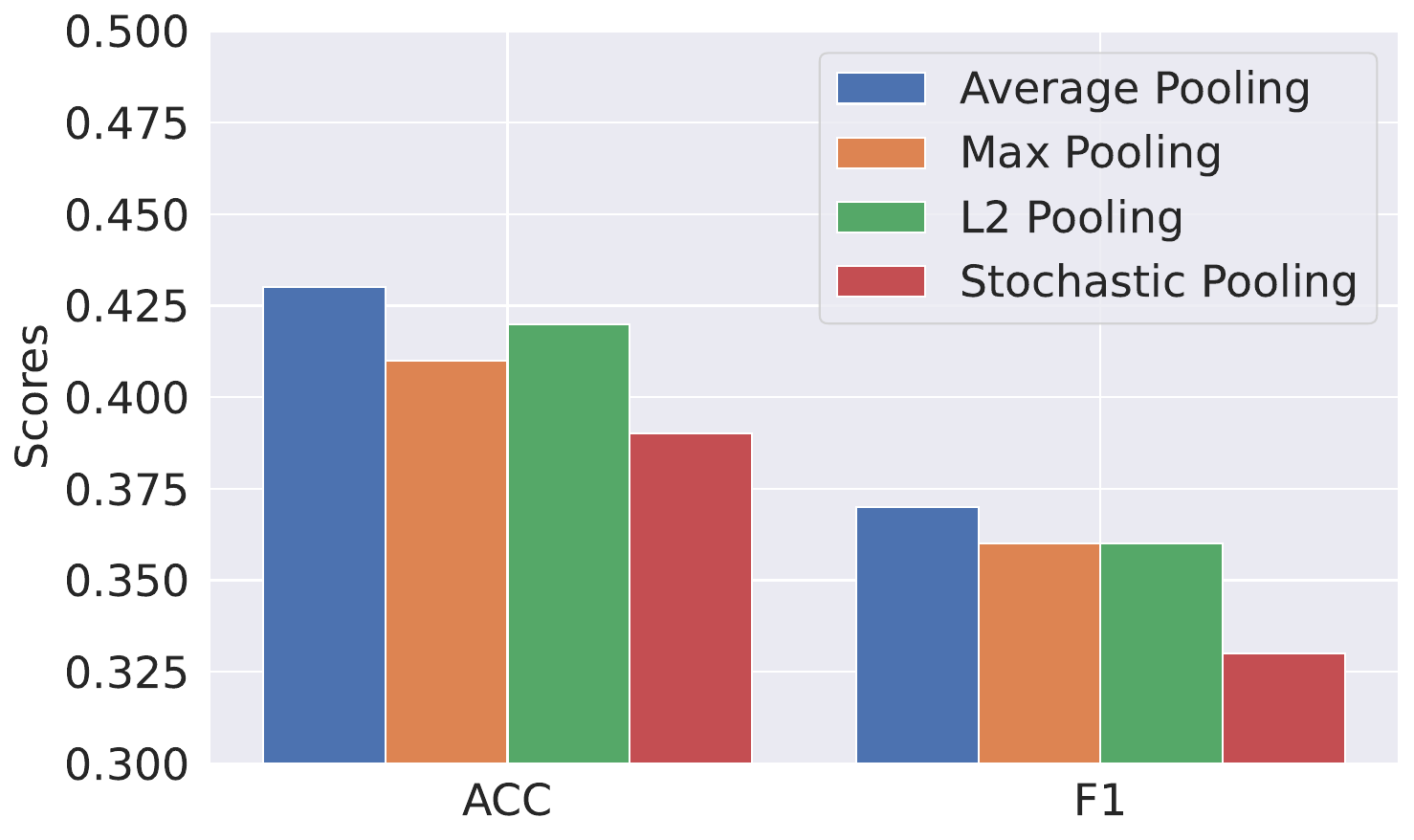}
        \caption{IP102}
    \end{subfigure}
    \caption{Impact of different pooling methods for selective weight generation in the selective module.}
    \label{fig:pooling}
\end{figure}
We investigate the impact of various pooling methods on the performance of the Selective Module within our InsectMamba model. Specifically, we investigated Average Pooling, Max Pooling, L2 Pooling, and Stochastic Pooling to synthesize the global features as prescribed in Equation~\ref{equ:pool}. Figure~\ref{fig:pooling} shows the comparative performance on two datasets, i.e., Farm Insects and IP102. For the Farm Insects dataset, Average Pooling achieved the best accuracy and F1 score, indicating its effectiveness in preserving feature representation for classification tasks. Moreover, the results on the IP102 dataset show a consistent trend. Average Pooling performs as well as in the Farm Insects dataset. 

\section{Conclusion}
In this work, we proposed a novel model, InsectMamba, for insect pest classification. The model is designed to amalgamate the strengths of State Space Models, Convolutional Neural Networks, Multi-Head Self-Attention mechanisms, and Multilayer Perceptrons. By integrating these varied visual encoding strategies through Mix-SSM blocks and a selective aggregation module, InsectMamba has showcased the capability to address the challenges posed by pest camouflage and species diversity. In the experiment, we conduct an extensive evaluation that compares our method and other strong competitors on five insect pest classification datasets. Experimental results show our model outperforms other models, which demonstrates the effectiveness of our model. We also illuminate the importance of each integrated module through comprehensive ablation studies.

\bibliographystyle{plainnat}
\bibliography{ref}
\appendix
\end{document}